\documentclass{article}

\usepackage[main, final]{neurips_2026}

\usepackage[utf8]{inputenc} 
\usepackage[T1]{fontenc}    
\usepackage{url}            
\usepackage{booktabs}       
\usepackage{amsfonts}       
\usepackage{nicefrac}       
\usepackage{microtype}      
\usepackage{xcolor}         
\usepackage{booktabs}
\usepackage{arydshln} 
\usepackage{multirow}
\usepackage{colortbl}
\usepackage{pifont}
\usepackage{enumitem}
\usepackage{booktabs}   
\usepackage{tabularx}   
\usepackage{colortbl}   
\usepackage{multirow}   
\usepackage{makecell}   
\usepackage{adjustbox} 
\usepackage{caption}
\usepackage{wrapfig}
\usepackage{graphicx}

\usepackage{tcolorbox}
\tcbuselibrary{skins, breakable}

\tcbset{
  aibox/.style={
    width=\linewidth,
    top=10pt,
    bottom=4pt,
    before skip=2pt,
    colback=gray!5!white,        
    colframe=gray!100!white,     
    boxrule=0.4pt,              
    coltitle=black,            
    colbacktitle=white,        
    enhanced,
    center,
    attach boxed title to top left={yshift=-0.1in,xshift=0.15in},
    boxed title style={
      boxrule=0pt,              
      colback=white,             
      colframe=white,            
      fontupper=\small\bfseries, 
      left=4pt, right=4pt,       
    },
  }
}

\newtcolorbox{AIbox}[2][]{aibox,title=#2,#1}

\newcommand{\model}{\textsc{AVIC}}
\newcommand{\fullmodel}{\textsc{Adaptive Visual Imagination Control}}

\newcommand{\modelR}{\textsc{AVIC-R}\xspace}

\usepackage{amsmath}
\usepackage{amssymb}
\usepackage{mathtools}
\usepackage{amsthm}
\usepackage{subcaption}
\usepackage{wrapfig}
\usepackage{xspace} 
\usepackage{enumitem}

\usepackage{hyperref}
\usepackage[capitalize,noabbrev]{cleveref}
\crefname{section}{Sec.}{Secs.}
\Crefname{section}{Section}{Sections}

\theoremstyle{plain}

\theoremstyle{definition}

\theoremstyle{remark}

\definecolor{darkgreen}{rgb}{0,0.5,0}
\definecolor{azureblue}{rgb}{0,0.5,1}
\definecolor{darkgreen}{rgb}{1,0,0}
\definecolor{color1}{HTML}{006EB8}
\definecolor{color2}{HTML}{009B55}

\usepackage{pifont}

\usepackage[capitalize]{cleveref}
\crefname{section}{Sec.}{Secs.}
\Crefname{section}{Section}{Sections}
\Crefname{table}{Table}{Tables}
\crefname{table}{Tab.}{Tabs.}
\crefname{appendix}{Sec.}{Secs.}
\Crefname{appendix}{Section}{Sections}

\usepackage{color,colortbl}
\definecolor{darkgreen}{rgb}{0,0.5,0}

\definecolor{darkgreen}{rgb}{1,0,0}

\definecolor{azureblue}{rgb}{0,0.5,1}

\title{When and How Much to Imagine:\\ Adaptive Test-Time Scaling with \\ World Models for Visual Spatial Reasoning}

\author{%
  Shoubin Yu$^{*1}$ \quad Yue Zhang$^{*1}$ \quad Zun Wang$^{1}$ \\ \textbf{Jaehong Yoon}$^{2}$ \quad \textbf{Huaxiu Yao}$^{1}$ \quad \textbf{Mingyu Ding}$^{1}$ \quad \textbf{Mohit Bansal}$^{1}$ \\ \\
  $^{1}$University of North Carolina, Chapel Hill \quad 
  $^{2}$Nanyang Technological University \\
  \\
  \url{https://adaptive-visual-tts.github.io}
}

\begin{document}

\maketitle

\begin{abstract}
Despite rapid progress in Multimodal Large Language Models (MLLMs), visual spatial reasoning remains unreliable when correct answers depend on how a scene would appear under unseen or alternative viewpoints. Recent work addresses this by augmenting reasoning with world models for visual imagination, but questions such as when imagination is actually necessary, how much of it is beneficial, and when it becomes harmful, remain poorly understood. In practice, indiscriminate imagination can increase computation and even degrade performance by introducing misleading evidence. 
In this work, we present an in-depth analysis of test-time visual imagination as a controllable resource for spatial reasoning. 
We first study when static visual evidence is sufficient, when imagination improves reasoning, and how unnecessary imagination affects accuracy and efficiency. 
To support this analysis, we then introduce \model, an adaptive test-time framework with world models that explicitly reasons about the sufficiency of current visual evidence before selectively invoking and scaling visual imagination. 
Finally, to further learn this gating and planning behavior without any annotation of when and how much to imagine, we introduce \modelR, which trains the policy end-to-end via GRPO from QA-correctness rewards and penalties by imagination cost.
Across spatial reasoning benchmarks (SAT, MMSI) and an embodied navigation benchmark (R2R), our results reveal clear scenarios where imagination is critical, marginal, or detrimental, and show that selective control can match or outperform fixed imagination strategies with substantially fewer world-model calls. Our \modelR surpasses strong proprietary baselines including GPT-4o and GPT-4.1 while invoking the world model less often. Overall, our findings highlight the importance of analyzing and controlling test-time imagination for efficient and reliable spatial reasoning.
\end{abstract}

\section{Introduction}
\label{sec_intro}
Recent advances in multimodal large language models (MLLMs)~\citep{li2024llava, li2023blip} have led to impressive progress in visual understanding and reasoning across various tasks.
These models can follow natural language instructions, perceive visual scenes, and reason over multimodal input to support decision making. Despite the progress, \emph{visual spatial reasoning} remains a persistent challenge~\citep{Yang2024ThinkingIS, Cheng2024SpatialRGPTGS, ray2024sat, Tong2024Cambrian1AF}, particularly for questions whose answer depends on unseen regions, viewpoint changes, or transformations that cannot be reliably inferred from a single static observation. 

\begin{figure*}[t!]
    \centering
    \includegraphics[width=0.98\linewidth]{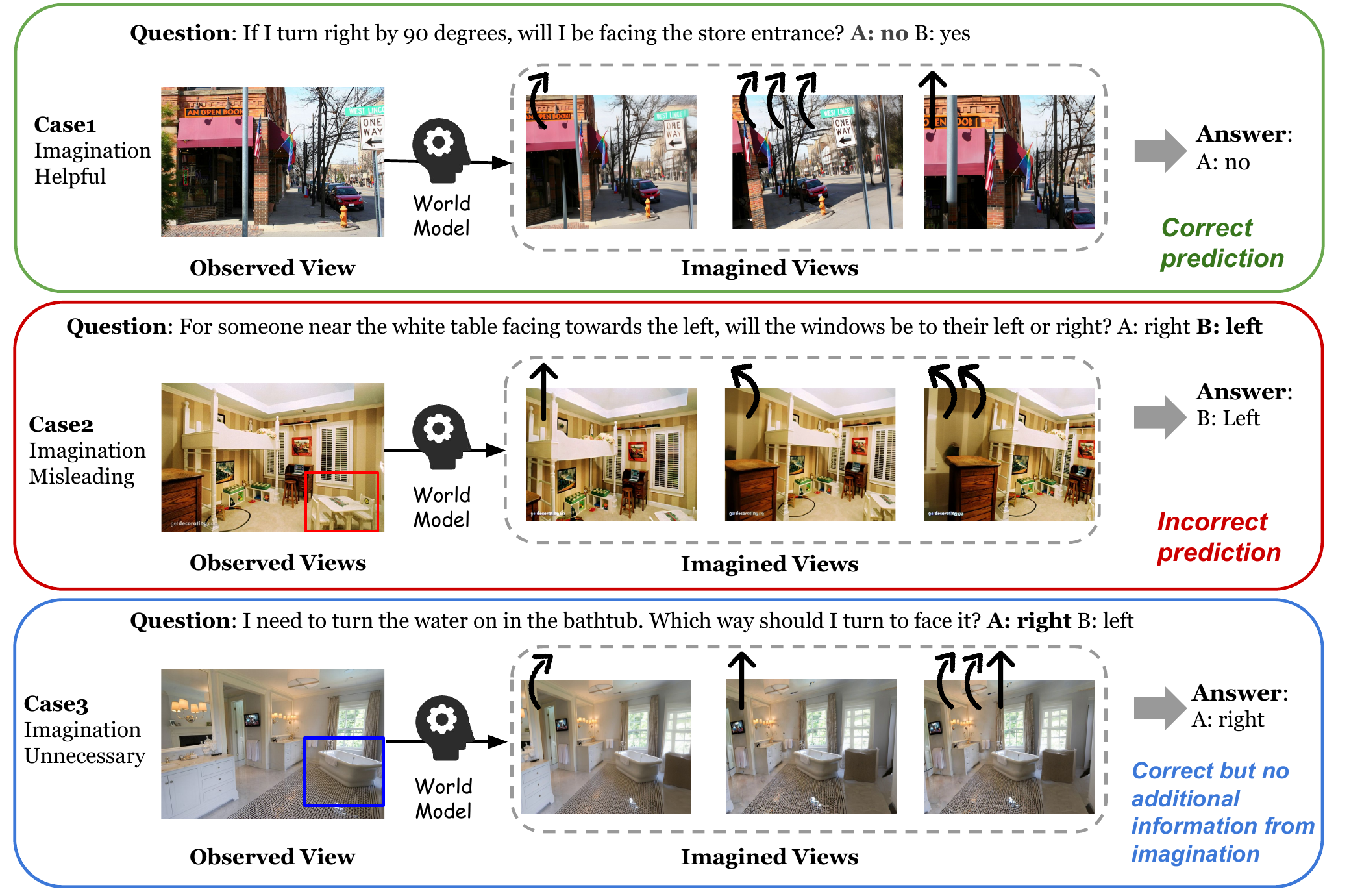}
    \caption{Different cases in always-on visual imagination.
Imagined views are generated independently for different beam-searched actions (shown by multiple arrows). Case 1 (Helpful): Visual imagination reveals previously unseen viewpoints, enabling helpful spatial reasoning.
Case 2 (Misleading imagined evidence): Imagination fails to preserve a task-relevant object (the white table in the red box). 
Case 3 (Unnecessary): The required information is already clearly observable in the original view (e.g., the bathtub in the blue box), making additional imagined views redundant. 
}
\vspace{-1mm}
    \label{fig:different casees}
\end{figure*}

A natural way to address this challenge, mirroring how humans operate, is through \emph{visual imagination}~\citep{kosslyn2006case}: when the observed visual evidence is insufficient, people mentally simulate how a scene would appear from alternative viewpoints or after potential movements, leveraging strong world priors learned from years of physical interaction and visual experience. 
Inspired by this intuition, recent work~\citep{yang2025mindjourney, cao2025spatialdreamer, qian2026current} has begun to integrate MLLMs with visual world models that can generate controlled novel views conditioned on hypothetical action at inference time.
However, existing approaches often invoke visual imagination using fixed and exhaustive strategies~(see \Cref{fig:different casees}), without first reasoning about whether additional imagination is necessary and helpful. 
This lack of deliberation can lead to problematic imagination, producing misleading (\Cref{fig:different casees} (b)) or redundant (\Cref{fig:different casees} (c)) views that not only incur substantial computational overhead but can also distract downstream reasoning and result in worse performance than relying on the original observation alone.
Through a systematic analysis of always-on imagination (\Cref{sec:analysis}), we show that such strategies are both inefficient and unreliable, motivating the need for more adaptive use of world models.

Based on these observations, we aim to answer two fundamental questions for visual spatial reasoning with world model imagination: \emph{when} should a model invoke visual imagination, and \emph{how much} imagined visual evidence is necessary if imagination is required.
Rather than treating visual imagination as an always-on operation, we seek to make it a controllable, self-adaptive component during inference time.
In this paper, we introduce \textbf{A}daptive \textbf{V}isual \textbf{I}magination \textbf{C}ontrol~(\textbf{\model}), a framework that gates and plans world-model usage with an explicit policy model.
Given an observation and a question, the policy first reasons about the sufficiency of the available visual evidence and conditionally decides whether to invoke the world model.
If it decides not to, it answers directly from the observed view; otherwise, it generates a dynamic-length action plan that specifies how the imagination should move or reorient to acquire informative viewpoints, which are rendered by the visual world model and consumed by a downstream reasoner.
A trajectory-level verifier then selects the most informative imagined trajectory among multiple policy samples, in contrast to prior beam-search approaches that score isolated keyframes.
It enables instance-dependent test-time scaling, allowing us to move beyond fixed or exhaustive imagination strategies and to study visual spatial reasoning systematically.

While \model~can be instantiated in a training-free manner via prompting and self-consistency sampling from a strong MLLM, training a stronger policy is non-trivial: no ground-truth supervision exists for what the optimal imagination trajectory should be, ruling out standard supervised fine-tuning or behavior cloning.
To overcome this challenge and learn this behavior end-to-end, we further propose \textbf{\modelR}, which trains the gating policy on top of small open-source model (Qwen2.5-VL~\citep{bai2025qwen2}) via reinforcement learning, using a composite reward built around QA correctness that requires no human supervision on \emph{when} imagination is necessary.
Specifically, we adopt Group-Relative Policy Optimization~(GRPO)~\citep{shao2024deepseekmath} to compute group-normalized advantages over rollouts that share the same prompt; the reward augments QA correctness with an action-count cost that discourages over-imagination, and a wrong-skip penalty that prevents the policy from collapsing to always-skip WM calling. These designs enable the policy model to discover better gating and planning behaviors from the QA accuracy and WM cost.

We evaluate the proposed framework on challenging spatial reasoning benchmarks (SAT~\cite{ray2024sat}, MMSI~\cite{yang2025mmsi}), and the navigation benchmark \textcolor{red}{}R2R~\cite{anderson2018vision}.
Across these settings, adaptive test-time scaling achieves SoTA or competitive performance while requiring substantially fewer extra language tokens and world-model calls compared to fixed imagination strategies.
Notably, with only a small training set and LoRA updates, \modelR~boosts a 7B open-source policy enough that the resulting pipeline outperforms variants using GPT-4o or GPT-4.1 as the policy model.
Overall, beyond improved performance, our results reveal that the benefits of visual imagination are highly instance-dependent and structured by the nature of the spatial reasoning query.
In particular, we find that world models are most beneficial for action-conditioned spatial reasoning, where answers depend on how a scene would evolve under specific movements or viewpoint changes, while offering limited gains for queries that can be resolved from existing observations.
At the same time, our analysis shows that effective visual spatial reasoning typically requires only targeted imagination, and excessive or indiscriminate simulation can introduce noise and degrade performance. Together, these findings indicate that visual imagination is a selective, query-dependent test-time resource, requiring adaptive, uncertainty-aware allocation of world-model computation.
\vspace{-3mm}
\section{Related Work}
\vspace{-3mm}
\noindent\textbf{Visual Spatial Reasoning with MLLMs.} 
The rapid evolution of Multimodal Large Language Models (MLLMs) has made significant progress in various downstream tasks~\cite{Li2022BLIPBL, Radford2021LearningTV, Liu2023VisualIT, Hong20233DLLMIT, zhang2024common,Yu2024CREMAGA,yu2025mexa, yu2025veggie, guo2025rethinking, yoon2024raccoon, deng2025motion}. 
In particular, spatial reasoning has attracted considerable attention due to its critical role in bridging visual perception with embodied tasks~\cite{Zhang2024VisionandLanguageNT, Wang2023ScalingDG,zhang2023vln, zhang2024spartun3d}. 
However, recent comprehensive evaluations indicate that current MLLMs still struggle with robust spatial reasoning~\cite{Yang2024ThinkingIS, Cheng2024SpatialRGPTGS, ray2024sat, Tong2024Cambrian1AF}. 
While recent efforts aim to enhance spatial capabilities through scaling training data~\cite{Chen2024SpatialVLMEV, Huang2023AnEG, li2024selma} or chain-of-thought prompting~\cite{Zhang2024ImproveVL, Ji2025EnhancingSR}, they fundamentally process visual information as static 2D snapshots. In contrast, robust spatial reasoning requires an active and dynamic process where the agent can selectively acquire new visual evidence, similar to human mental simulation.

\noindent\textbf{World Models and Visual Imagination.} 
Recent advances in video generation have demonstrated the potential of serving as world models, enabling agents to imagine future frames or outcomes for improved decision-making~\cite{Zhou2018StereoM, Du2023LearningUP, Li2023PanoGenTP, huang2025planning,qian2026current}.
This capability is further boosted by the emergence of controllable video generation, which allows for action-conditioned simulation~\cite{He2024CameraCtrlEC, Bahmani2024AC3DAA, Yu2024ViewCrafterTV, Wang2025EPiCEV}. 
Notably, recent works such as MindJourney~\cite{yang2025mindjourney} have pioneered the use of world models to enhance visual spatial reasoning by synthesizing novel viewpoints. 
However, their model blindly generates a set number of views regardless of the question's difficulty or necessity. In contrast, we show that the utility of visual imagination is highly query-dependent, motivating selective rather than uniform use of world models at test time.

\noindent\textbf{Test-Time Scaling.}
Test-time scaling (TTS) improves performance by allocating additional inference computation without retraining. Prior work has explored various scaling strategies in language/visual-language models, including self-consistency~\cite{wang2022self}, tree-based search~\cite{yao2023tree,wang2025video}, verifier-guided method~\cite{lifshitz2025multi, zhang2024generative}, and (multimodal) CoT~\cite{yan2025videochat, xu2025softcot++, wang2025multimodal}. In the visual spatial reasoning, recent works~\cite{yang2025mindjourney, cao2025spatialdreamer} realized through generating novel views and ensembling, but typically apply uniform computation across instances. Our method introduces adaptive visual test-time scaling, enabling targeted imagination only when necessary and improving computational efficiency.
\begin{figure*}[t]
  \centering
  \begin{subfigure}[t]{0.32\linewidth}
    \centering
    \includegraphics[width=\linewidth]{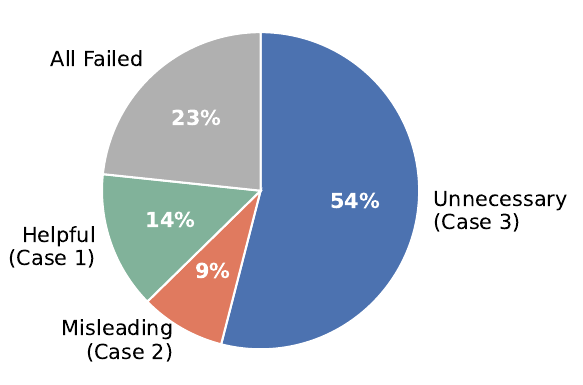}
    \caption{Statistics of Different Cases of Always-on Imagination}
    \label{fig:distribution}
  \end{subfigure}\hfill
  \begin{subfigure}[t]{0.31\linewidth}
    \centering
    \includegraphics[width=\linewidth]{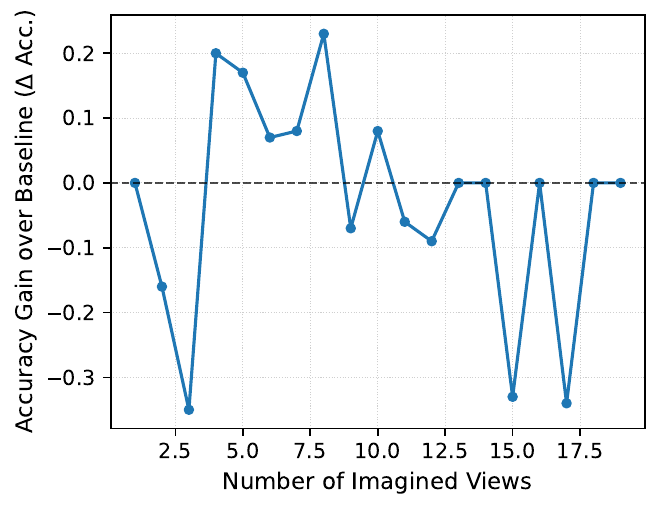}
    \caption{Performance Gain v.s. Amount of Imagined Views}
    \label{accuracy_gain_vs_views}
  \end{subfigure}\hfill
  \begin{subfigure}[t]{0.32\linewidth}
    \centering
    \includegraphics[width=\linewidth]{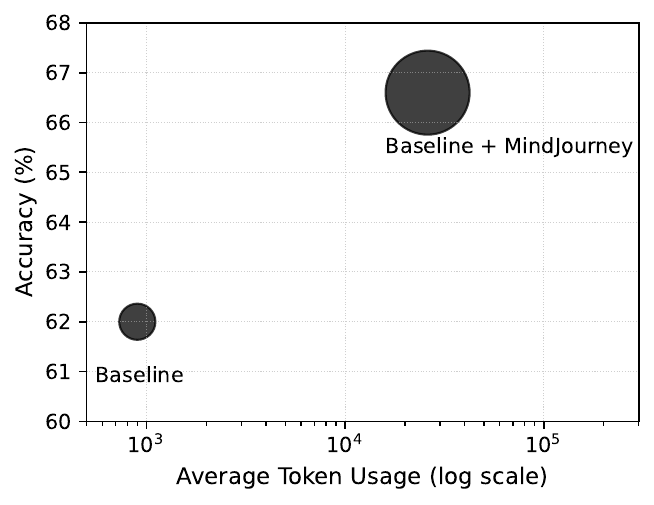}
    \caption{Trade-off Among Accuracy, Token Numbers and Inference Time}
    \label{acc_token_wm_bubble}
  \end{subfigure}

  \caption{\textbf{(a)}: In the majority of cases, visual imagination is unnecessary, while a smaller fraction is helpful or misleading, highlighting the need for selective invocation rather than uniform use.
\textbf{(b)}: Accuracy gain over the baseline over the number of imagined views. Performance improvements are non-monotonic, indicating that additional imagination does not consistently translate to better reasoning and may even degrade accuracy when there are too many generated views.
\textbf{(c)}: Accuracy v.s. average token usage. Bubble size indicates average running time. Fixed imagination strategies achieve higher accuracy at the cost of substantially increased computation, motivating adaptive test-time scaling that balances performance and efficiency.}
  \label{fig:analyze}
\end{figure*}

\section{Analysis of Always-on World Model Calling}
\label{sec:analysis}
\label{background:world model}
We consider a test-time setting where an MLLM is equipped with a visual world model that generates imagined observations from hypothetical viewpoints. Existing methods commonly invoke the world model in an always-on, which calling it on every instance and exhaustively exploring action branches, implicitly assuming that additional imagination is consistently beneficial. In practice, however, this incurs substantial computational cost and may yield ambiguous or noisy observations, making imagined views redundant when the answer is already evident from the initial observation and outright misleading when the world model produces noise.

\label{empirical statistics}
To diagnose this strategy, we categorize each instance on SAT-Real~\cite{ray2024sat} into three cases (\Cref{fig:different casees}):

\begin{itemize}[leftmargin=*, itemsep=2pt, topsep=2pt]
    \item \textbf{Case 1 (Imagination Helpful):} The model calls the world model and produces a correct answer, indicating that imagined views provide beneficial spatial information.
    \item \textbf{Case 2 (Misleading Imagined Evidence):} The model produces an incorrect answer after using imagined views. This outcome can reflect rendering errors, uninformative trajectories, or failure to use the imagined evidence.
    \item \textbf{Case 3 (Imagination Unnecessary):} The model produces a correct answer without calling the world model, suggesting that visual imagination is redundant.
\end{itemize}

We then examine three aspects of always-on behavior in \Cref{fig:analyze}: case distribution, performance vs.\ amount of imagination, and computational cost.
\noindent\textbf{(1) Case distribution.} As shown in \Cref{fig:distribution}, the majority of instances (54\%) fall into Case~3, where the model already answers correctly without any world model invocation, while imagination is genuinely helpful in only 14\% (Case~1), indicating that always-on imagination is unnecessary for most instances.
\noindent\textbf{(2) Performance vs.\ amount of imagination.} \Cref{accuracy_gain_vs_views} shows that adding more imagined views does not consistently improve accuracy and even degrades it, suggesting that simply increasing imagination is not an ideal strategy.
\noindent\textbf{(3) Cost--accuracy trade-off.} While always-on imagination yields only a 4.6\% accuracy gain over the baseline, it requires nearly two orders of magnitude more tokens and about $30\times$ higher inference time (\Cref{acc_token_wm_bubble}), a steep computational price for limited return.
\noindent\textbf{(4) Selective imagination upper bound.} We further quantify the potential of selective WM usage by assuming imagination is applied only when it leads to a correct prediction. The baseline reaches \textbf{62.0\%} on SAT-Real and always-on imagination only marginally improves to \textbf{66.6\%}, whereas this selective upper bound jumps to \textbf{75.3\%}, demonstrating that selective imagination policies are strongly motivated.
The 75.3\% selective upper bound applies to the fixed imagination procedure in this diagnostic, rather than to all possible trajectories or to \modelR; Appendix~\ref{app:oracles} reports oracle interventions for \modelR. These results motivate selective, adaptive imagination.

\section{Adaptive Visual Imagination Control}
\label{sec:method}
\begin{figure*}
    \centering
    \includegraphics[width=\linewidth]{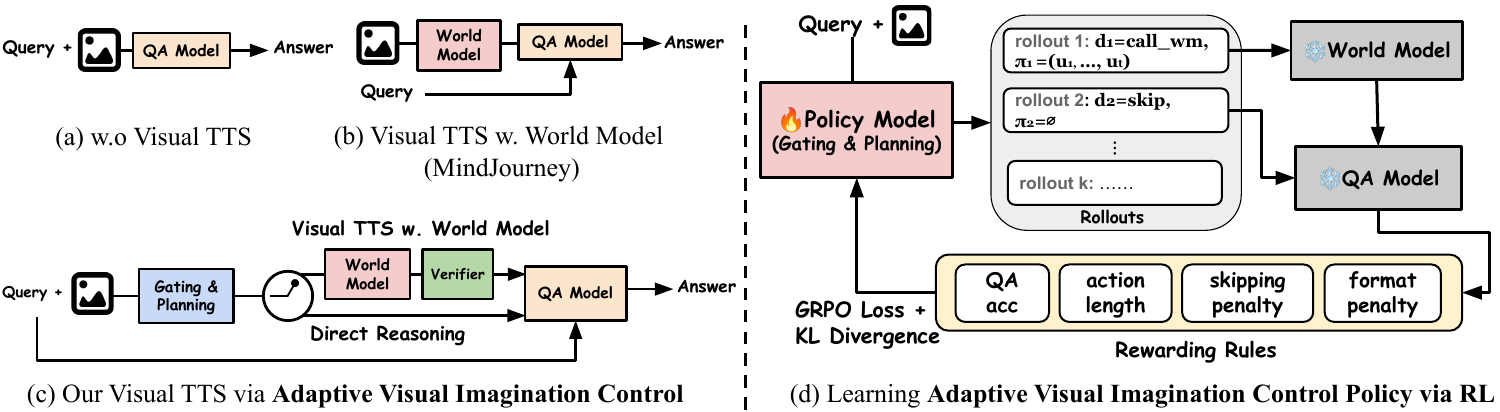}
    \caption{(a) Direct QA from the current observation. (b) Always-on world-model exploration. (c) Ours: a policy model decides \emph{whether} and \emph{how much} to imagine, selectively querying the world model only when warranted. (d) Our RL training loop for the gating policy: The policy model samples a set of rollouts that either query the world model (\texttt{call\_wm}) or bypass it (\texttt{skip}). A frozen QA model answers them, and a four-component reward drives a GRPO update on the policy model.
    }
    \label{fig:method}
\end{figure*}

We now introduce \textbf{\model~(\fullmodel)}, an adaptive test-time framework that selectively invokes a world model only when additional visual evidence is likely to be useful (\cref{fig:method}c), in contrast to the always-on baseline analyzed in~\cref{sec:analysis}. We first formalize the problem (\cref{sec:problem}), then describe the framework (\cref{sec:framework}), and finally describe an RL procedure that trains the gating policy from QA Model and WM rewards (\cref{sec:rl}).

\subsection{Problem Formulation}
\label{sec:problem}
We consider a visual spatial reasoning task defined by an input tuple $\langle I, q, \mathcal{A}\rangle$, where $I$ is the current egocentric observation (one or multiple views), $q$ is a multiple-choice question, and $\mathcal{A}=\{a_1,\dots,a_K\}$ is the answer set. The correct answer may depend on spatial relations that are ambiguous, occluded, or unobserved in $I$. The agent may optionally invoke a visual world model that renders novel imagined observations $I_{\pi}$ from a sequence of egocentric actions $\pi$, with $I_{\pi}=\varnothing$ when no imagination is used. The predicted answer is
$\hat{a} = \arg\max_{a \in \mathcal{A}} P\big(a \mid I, I_{\pi}, q\big).
$

\subsection{Adaptive Imagination Framework}
\label{sec:framework}
\model~couples a \textbf{policy model} that gates and plans imagination with a \textbf{trajectory-level verifier} that picks a single targeted imagined trajectory for downstream reasoning.

\noindent\textbf{Policy gating with test-time scaling.} A policy $\theta$ maps $(I, q, \mathcal{A})$ to a decision $d \in \{\texttt{skip}, \texttt{call\_wm}\}$ together with a short discrete action plan $\pi$ drawn from a fixed low-level egocentric action space $\mathcal{U}$:
\begin{equation}
(d, \pi) \sim \theta(d, \pi \mid I, q, \mathcal{A}),\quad
\pi = \begin{cases}\varnothing, & d=\texttt{skip},\\ (u_1,\dots,u_T),\ u_t \in \mathcal{U}, & d=\texttt{call\_wm}.\end{cases}
\label{eq:joint_policy}
\end{equation}
Here $\mathcal{U}$ contains \texttt{turn-left} ($9^\circ$), \texttt{turn-right} ($9^\circ$), and \texttt{move-forward} ($0.25$\,m), following MindJourney. A plan contains at most six atomic actions; repeated primitives compose larger motions, with cancelling or oscillatory sequences disallowed. Plans are converted to relative camera poses for world-model conditioning.
We sample the policy $M$ times ($M=5$ by default) and apply majority voting only to the gating decision $d$. This vote determines whether to imagine but not select a plan.

\noindent\textbf{Action execution and trajectory selection.} When the majority decision is $\texttt{call\_wm}$, the world model renders the non-empty candidate plans from the policy samples, producing $\mathcal{I}_{\pi^{(m)}} = W(I, \pi^{(m)})$. The verifier scores each complete rendered trajectory, and only the highest-scoring trajectory is passed to the reasoner; it is not applied only to one preselected plan. Different policy samples produce trajectories of varying usefulness; unlike prior beam-search approaches~\cite{yang2025mindjourney} that score isolated keyframes, we evaluate the \emph{entire} trajectory as a coherent unit via a verifier $V$, preserving temporal and geometric consistency across sequential actions. Such that:
\begin{equation}
s^{(m)} = V\!\left(I, q, \mathcal{I}_{\pi^{(m)}}\right),\quad \pi^{*} = \arg\max_{\pi^{(m)}} s^{(m)},
\label{eq:traj_select}
\end{equation}

\noindent\textbf{Final prediction.} A vision-language reasoner $\phi$ predicts the answer using the original observation and the selected imagined views (or $I$ alone when the gate selected $\texttt{skip}$, in which case $I_{\pi^{*}}=\varnothing$):
\begin{equation}
\hat{a} = \arg\max_{a \in \mathcal{A}} P_\phi\!\left(a \mid I, I_{\pi^{*}}, q\right).
\label{eq:final_pred}
\end{equation}

\subsection{Learning When and How Much to Imagine via RL}
\label{sec:rl}

Learning when and how much to imagine is non-trivial, as no ground-truth supervision exists for what the optimal imagination trajectory should be, ruling out supervised fine-tuning or behavior cloning. We therefore propose \modelR, which trains $\theta$ end-to-end with reinforcement learning, using QA correctness as the primary signal augmented by lightweight reward shaping (\Cref{fig:method} (d)).

\noindent\textbf{Training loop.} For each question, the policy $\theta$ samples $K$ rollouts $\{(d^{(i)}, \pi^{(i)})\}$. Each rollout is executed by frozen environment modules: $\texttt{call\_wm}$ rollouts query the world model $W$ to render imagined views, while $\texttt{skip}$ rollouts bypass it; both are answered by a frozen QA model $\phi$.
The resulting reward signals are aggregated into a GRPO update on the policy, while $W$ and $\phi$ remain frozen throughout training. The verifier is used only at inference. QA correctness is computed by comparing the frozen QA model\textquotesingle s predicted answer with the dataset ground-truth label. We also discussed dependence on the training reasoner in Appendix~\ref{app:transfer}.

\noindent\textbf{Reward design.} We assemble a composite reward with 4 components, as illustrated in \Cref{fig:method} (d):
\begin{equation}
r \;=\; \underbrace{\mathbf{1}_{\text{correct}}}_{\text{(i) QA correctness}}
\;-\; \underbrace{c\,|\pi|}_{\text{(ii) action length}}
\;-\; \underbrace{\beta_s\,\mathbf{1}_{\text{wrong-skip}}}_{\text{(iii) wrong-skip}}
\;-\; \underbrace{\beta_p\,\mathbf{1}_{\text{parse-fail}}}_{\text{(iv) format}}
\label{eq:reward}
\end{equation}
where $\mathbf{1}_{\text{correct}}$, $\mathbf{1}_{\text{wrong-skip}}$, and $\mathbf{1}_{\text{parse-fail}}$ indicate, respectively, that $\hat a = a^*$, that $d{=}\texttt{skip} \land \hat a {\ne} a^*$, and that the output schema cannot be parsed; $|\pi|$ counts atomic actions; and $c, \beta_s, \beta_p > 0$. Each term targets a specific failure mode:
\begin{itemize}[leftmargin=*, itemsep=2pt, topsep=2pt]
\item \textbf{QA correctness} is the only positive signal; all other terms are deductions.
\item \textbf{Action-length cost} discourages over-imagination, a longer correct chain still scores below a shorter correct one, pushing the policy toward concise plans.
\item \textbf{Wrong-skip penalty} is essential. Without it, an incorrect $\texttt{skip}$ pays nothing while an incorrect $\texttt{call\_wm}$ still pays $c|\pi|$, biasing the policy toward skipping under uncertainty.
\item \textbf{Format penalty} handles unrecoverable schema errors. We deliberately keep $\beta_{\text{p}}\!\approx\!\beta_{\text{s}}$ rather than larger: a heavier format penalty would incentivize the policy to collapse onto trivial outputs (e.g., always $\texttt{skip}$ with empty $\pi$) merely to avoid parse failures.
\end{itemize}
We set $c=0.1$ and $\beta_{\text{s}}=\beta_{\text{p}}=0.5$, yielding a clean qualitative ordering: a correct $\texttt{skip}$ scores $+1.0$, a correct $\texttt{call\_wm}$ with $n$ actions scores $1-0.1n$, and a wrong $\texttt{skip}$ ($-0.5$) is strictly worse than a short wrong $\texttt{call\_wm}$ ($-0.1$ to $-0.3$). The goal of this asymmetry is to prevent the policy from collapsing to \emph{always-skip} under uncertainty: when in doubt, calling WM with a short plan is the safer bet than skipping. The action cost $c|\pi|$ then discourages over-long imagination chains.

\paragraph{GRPO objective.} Without an external value function, we adopt Group-Relative Policy Optimization~\cite{shao2024deepseekmath}, which estimates advantages directly from the $K$ rollouts sharing a prompt:
\begin{equation}
A_i = \frac{r_i - \mu_q}{\sigma_q + \epsilon},\quad
\mu_q = \tfrac{1}{K}\!\sum_{j} r_j,\quad \sigma_q^{2} = \tfrac{1}{K}\!\sum_{j}(r_j - \mu_q)^{2}.
\label{eq:advantage}
\end{equation}
Group normalization absorbs per-question difficulty, where uniformly easy or uniformly hard groups contribute zero gradient. We optimize a token-level PPO-clipped objective~\cite{schulman2017proximal} regularized by KL to a frozen reference $\theta_{\text{ref}}$:
\begin{equation}
\mathcal{L} = -\mathbb{E}_{(q,i,t)}\!\left[\min\!\big(\rho_{i,t}A_i,\;\mathrm{clip}(\rho_{i,t},1{-}\epsilon_c,1{+}\epsilon_c)A_i\big)\right] + \beta_{\text{KL}}\,\widehat{\mathrm{KL}}\!\left(\theta\|\theta_{\text{ref}}\right),
\label{eq:grpo_loss}
\end{equation}
with per-token importance ratio $\rho_{i,t} = \theta(y_{i,t}\mid q,y_{i,<t}) / \theta_{\text{old}}(y_{i,t}\mid q,y_{i,<t})$. Only LoRA adapters in $\theta$ are updated; the KL anchor preserves the base VLM's general capabilities while shaping only gating and planning behaviors. Training hyperparameters, data curation, output parsing, and training-compute accounting are detailed in Appendix~\ref{app:implementation}.

\begin{table*}[t]
\renewcommand\arraystretch{1.1}
\setlength{\tabcolsep}{8pt} 
    \caption{Comparison between TTS methods on SAT-Real. The best results are denoted by \textbf{bold}, and the second-best are \underline{underlined}. \textbf{Avg. WM}: average world model calling times over the dataset.}
    \label{table:SAT}
    \centering
    \resizebox{0.9\linewidth}{!}{
\begin{tabular}{lcccccccrr}
\toprule
\textbf{Method}  & \multicolumn{1}{c}{\textbf{Policy Model}} &  \multicolumn{1}{c}{\textbf{EgoM}} & \multicolumn{1}{c}{\textbf{ObjM}} & \multicolumn{1}{c}{\textbf{EgoAct}} & \multicolumn{1}{c}{\textbf{Goal}} & \multicolumn{1}{c}{\textbf{Pers}} & \multicolumn{1}{c}{\textbf{Avg.}} & \multicolumn{1}{c}{\textbf{\# Token (K)}} & \multicolumn{1}{c}{\textbf{Avg. WM}}  \\ \midrule
InternVL3-14B~\cite{zhu2025internvl3} & -- & 56.5 & \underline{69.5} & 54.0 & 73.5 & \underline{45.4} & 59.3  & 0.2 & 0\\
+ MindJourney & -- & 69.6 & 60.9 & \textbf{78.4} & \underline{79.4}  & 42.4 &  66.7 & 2.5 & 12.34 \\
+ \model & InternVL3-14B & \textbf{95.6} & \textbf{73.9} & 62.1 & 76.4 & 42.4 & \underline{68.0} & 2.0 & 0.64 \\
+ \model & Qwen2.5VL-7B & 73.9 & 47.8 & 67.5 & 73.5 & 42.4 & 61.3  & 4.4 & 1.81 \\
+ \modelR & Qwen2.5VL-7B &  \underline{82.6} & 52.1 & \underline{70.2} & \textbf{85.2} & \textbf{54.5} & \textbf{69.3} &4.8& 3.03\\ \midrule
GPT-4o~\cite{gpt4o} & -- & 56.5 & \textbf{85.0} & 50.0 & 64.0 & 45.0 & 60.3 & 0.9 & 0 \\
+ MindJourney &  -- & 78.3 &  60.9 &  \underline{78.4} & 70.6 &  \underline{57.5} & 69.3 & 26.0 & 12.34 \\
+ \model & GPT-4o & \textbf{86.9} & 60.9 & 64.8 &  \underline{82.3} & 48.4 & 69.3 & 9.5 & 0.72 \\
+ \model & Qwen2.5VL-7B & 65.2 & 73.9 & 64.8 & \textbf{91.1} & \textbf{60.6} & \underline{71.3}  & 5.0 & 1.81 \\
+ \modelR & Qwen2.5VL-7B & \underline{82.6} & \underline{82.6} & \textbf{81.0} & \textbf{91.1} & 51.2 & \textbf{77.3} & 5.4 & 3.03\\ \midrule
GPT-4.1~\cite{openai2024gpt4_1} & --  & \underline{95.7} & 73.9 & 78.3 & \textbf{88.2} & 39.4 & 74.0 & 0.7 & 0 \\
+ MindJourney & -- & \textbf{100.0}  & \underline{82.6} & \textbf{86.5} & 79.4 & 45.4 & 77.3 & 67.1 &12.34  \\
+ \model & GPT-4.1 & \textbf{100.0} & 78.2 & \underline{83.7}  & \underline{85.2} & \underline{54.5} & \underline{79.3} & 7.6  & 0.73 \\
+ \model & Qwen2.5VL-7B &  82.6 & \textbf{86.9} & 75.6 & \textbf{88.2} & 36.3 & 72.6 & 4.8 & 1.81\\
+ \modelR & Qwen2.5VL-7B & 91.3 & \textbf{86.9} & \underline{83.7} & \underline{85.2} & \textbf{57.5} & \textbf{80.0} &5.2 & 3.03\\ \midrule
o1~\cite{jaech2024openai} & -- & 78.3 & \underline{82.6} & 73.0 & 73.5 & \textbf{69.7} & 74.6 & 1.4 & 0  \\
+ MindJourney & -- & \textbf{100.0} & 65.2 & 78.4 & 82.4 & 63.7 & 77.3 &  39.4  &  12.34 \\
+ \model & o1 & \textbf{100.0} & \textbf{86.9} & \textbf{86.4} & \underline{91.1} & 66.6 & \textbf{85.3} & 14.6 &  1.28 \\
+ \model & Qwen2.5VL-7B & \underline{86.9} & 65.2 & 78.3 & \textbf{94.1} & \underline{69.6} & 79.3 & 5.7 & 1.81 \\
+ \modelR & Qwen2.5VL-7B & \underline{86.9} & 65.2 & \underline{81.0} & \textbf{94.1} & \underline{69.6} & \underline{80.0} & 6.1 & 3.03\\ \midrule
\end{tabular}
}
\end{table*}

\section{Experiments}
\label{sec:experiments}

\begin{table}[t]
\centering
\footnotesize
\setlength{\tabcolsep}{4pt}
\renewcommand{\arraystretch}{1.1}

\begin{minipage}[t]{0.36\columnwidth}
\centering
\caption{Results on MMSI.}
\label{tab:mmsi}
\begin{tabular}{lr}
\toprule
\textbf{Method} & \textbf{Accuracy} \\
\midrule
GPT-4o~\cite{gpt4o} & 30.3 \\
GPT-4o + \model & \textbf{32.3} \\ \midrule
GPT-4.1~\cite{openai2024gpt4_1} & 30.9 \\
GPT-4.1 + \model & \textbf{33.8} \\
\bottomrule
\end{tabular}
\end{minipage}
\hfill
\begin{minipage}[t]{0.62\columnwidth}
\centering
\caption{Results on R2R embodied navigation dataset.}
\label{tab:r2r}
\begin{tabular}{lccccc}
\toprule
\textbf{Methods} & \textbf{LLMs} & \textbf{NE$\downarrow$} & \textbf{OSR$\uparrow$} & \textbf{SR$\uparrow$} & \textbf{SPL$\uparrow$}  \\
\midrule
NavGPT~\cite{zhou2024navgpt} & GPT-3.5 & 8.02 & 26.4 & 16.7 & 13.0 \\
MapGPT~\cite{zhang2024mapgpt} & GPT-4 & 5.80 & 61.6 & 41.2 & 25.4 \\ \midrule
MapGPT & GPT-4o & 6.04 & 41.6 & 36.0 & 30.8 \\ 
MapGPT + \model & GPT-4o & \textbf{5.97} & \textbf{45.3} & \textbf{37.5} & \textbf{31.9}  \\ 
\bottomrule
\end{tabular}
\end{minipage}

\end{table}

\subsection{Experiment Setup}
\label{sec:setup}

\noindent \textbf{Datasets and Benchmarks.}
\label{sec:datasets}
We validate our proposed framework on both visual spatial reasoning benchmarks and embodied navigation tasks, covering a range of spatial ambiguities and interaction requirements.
For visual spatial reasoning, we evaluate on SAT~\cite{ray2024sat} and MMSI~\cite{yang2025mmsi}, two benchmarks designed to test visual spatial reasoning with single/multiple images. 
We also evaluate on the Room-to-Room (R2R)~\cite{anderson2018vision} for the embodied navigation task.

\noindent \textbf{Implementation Details.}
Our framework is implemented on top of a vision-language model and a pretrained visual world model, Stable Virtual Camera (SVC)~\cite{zhou2025stable}. 
In the default same-backbone~\model~setting, the policy, verifier, and final QA model use the same base MLLM with different prompts.
All decisions by AVIC are made at test time without additional fine-tuning, while we train AVIC-R with LoRA (More details in Appendix~\ref{app:implementation}).

\subsection{Main Results}
\textbf{\model~beats the always-on baseline in both efficiency and effectiveness across all backbones.} \Cref{table:SAT} compares our framework against baselines on SAT-Real~\cite{ray2024sat} across five categories. Across all open-source and proprietary backbones, \model~consistently improves over the base MLLM and matches or surpasses the always-on baseline (MindJourney) while using $\sim$10\% of the tokens and far fewer world-model calls. With GPT-4.1, accuracy rises from 74.0\% to 79.3\%; with o1, it reaches \textbf{85.3\%} (+10.7\% over base). 
Gains are most pronounced on Egocentric Movement, Action Consequence, and Perspective tasks, categories requiring action-conditioned spatial reasoning where selective imagination is most beneficial. \Cref{tab:mmsi} confirms these gains transfer to MMSI-Bench~\cite{yang2025mmsi}.

\textbf{\modelR turns a small policy model much stronger, even better than proprietary ones.
}
\modelR~adds two further findings on top of \model. First, a small RL-trained policy can drive a much larger backbone: with only Qwen2.5VL-7B as the gating policy, \modelR~outperforms \model~variants that use the proprietary backbone itself as the policy on three of four backbones. 
As shown in~\cref{table:SAT}, with GPT-4o used as the QA model, \modelR with Qwen2.5VL-7B outperforms GPT-4o used as the policy model by 8.0\% in QA accuracy, suggesting that \modelR enables better gating and planning of the world model. 
Second, RL is essential to making the small policy work: prompting Qwen2.5VL-7B alone as the \model~policy often underperforms even the always-on baseline (e.g., 61.3\% vs.\ MindJourney's 66.7\% on InternVL3-14B; 72.6\% vs.\ 77.3\% on GPT-4.1), as the 7B model lacks the in-context reasoning to gate reliably. Our lightweight RL training fixes this, lifting the same Qwen2.5VL-7B by 6--8 points across InternVL3/GPT-4o/GPT-4.1 and turning a policy that previously underperformed the always-on baseline into one that drives the full pipeline to top results. 

\textbf{\model's selective imagination also transfers to embodied navigation.} \Cref{tab:r2r} applies \model~to embodied navigation, integrated into MapGPT's~\cite{zhang2024mapgpt} step-wise framework on the 72-scene R2R~\cite{anderson2018vision} evaluation. 
At each step, our policy decides whether to augment currently observed panoramas with imagined views before the next-action prediction (more details in Appendix~\ref{app:interpretation}).
Compared to MapGPT with GPT-4o, \model~achieves higher OSR/SR/SPL and lower navigation error (NE), indicating more reliable goal reaching with shorter, less redundant trajectories, transferring the benefits of selective imagination from static spatial reasoning to embodied tasks.

\begin{table}[t]
\centering
\small
\setlength{\tabcolsep}{4pt}
\begin{minipage}[t]{0.58\linewidth}
\centering
\caption{Ablation over action scaling, gating, and world model. Based on GPT-4.1.}
\label{tab:ablation}
\begin{tabular}{ccccc}
\toprule
\textbf{Action Scaling} & \textbf{Gating} & \textbf{WM} & \textbf{Avg. WM} & \textbf{Acc.\ (\%)} \\
\midrule
--           & --           & --           & 0     & 74.0 \\
--           & --           & $\checkmark$ & 12.34 & 77.3 \\
--           & $\checkmark$ & $\checkmark$ & 0.51  & 73.3 \\
$\checkmark$ & $\checkmark$ & $\checkmark$ & 0.73  & 79.3 \\
\bottomrule
\end{tabular}
\end{minipage}
\hfill
\begin{minipage}[t]{0.35\linewidth}
    \centering
    \caption{Effect of policy and QA model choice on SAT-Real.}
    \label{tab:policy_qa}
    \begin{tabular}{llc}
        \toprule
        \textbf{Policy} & \textbf{QA} & \textbf{Acc.\ (\%)} \\
        \midrule
        GPT-4o & GPT-4o & 68.0 \\
        GPT-4o & o1     & 80.0 \\
        o1     & o1     & 81.3 \\
        o1     & GPT-4o & 68.6 \\
        \bottomrule
    \end{tabular}
\end{minipage}
\end{table}

\vspace{-3mm}
\begin{table}[]
\centering
\caption{\textbf{Reward ablation: the \emph{skip-wrong} penalty.} Removing the $-0.5$ penalty for an incorrect \texttt{skip} causes excessive skipping, dropping overall accuracy by $14.66$ points and degrading every question type. Values are test accuracy (\%).}
\label{tab:ablation_reward}
\small
\begin{tabular}{l c c c c c c}
\toprule
 & EgoM & ObjM & Goal & EgoAct & Pers & \textbf{Avg.} \\
\midrule
\modelR \emph{w/o} skip-wrong & $65.22$ & $65.22$ & $79.41$ & $62.16$ & $42.42$ & $62.67$ \\
\modelR (full)                & $\mathbf{82.61}$ & $\mathbf{82.61}$ & $\mathbf{91.18}$ & $\mathbf{81.08}$ & $\mathbf{51.52}$ & $\mathbf{77.33}$ \\
\midrule
$\Delta$ & $+17.39$ & $+17.39$ & $+11.77$ & $+18.92$ & $+9.10$ & $+14.66$ \\
\bottomrule
\end{tabular}
\end{table}

\subsection{Ablation Studies}
\noindent\textbf{Effect of selective gating and action-level scaling.} 
As listed in~\cref{tab:ablation}, we analyze the contributions of world-model (WM) imagination, gating, and action-level test-time scaling. Vanilla baseline achieves 74.0\% w.o any scaling. Always-on invoking the WM with spatial beam search and without gating or action scaling improves performance to 77.3\%, but at the cost of excessive computation, requiring an average of 12.34 WM calls.
Introducing a gating mechanism via a policy model alone drastically reduces WM usage (0.51 calls) but also hurts accuracy (73.3\%), indicating that binary WM invocation without action-level control is insufficient and can suppress necessary imagination.
In contrast, our full method that combines gating with action scaling achieves the best performance (79.3\%) while keeping WM usage low (0.73 calls).
This demonstrates that \emph{when} to invoke the WM and \emph{how} to use it are both critical: selective gating must be paired with targeted action planning to ensure that limited imagination is informative rather than restrictive.
Overall, it highlights that effective visual TTS requires control over both WM invocation and action planning.

\noindent\textbf{Effect of policy and QA model choice.}
As listed in~\cref{tab:policy_qa}, we compare different combinations of policy models and QA models on SAT-Real.
We find upgrading the QA model yields substantial improvements regardless of the policy model used (68.0\% $\rightarrow$ 80.0\%).
These comparisons show that the choice of QA model affects accuracy, but they do not uniquely identify the system bottleneck. The separate selection and oracle studies in Appendices~\ref{app:selection} and~\ref{app:oracles} show additional headroom from gating and trajectory selection with the reasoner held fixed.

\noindent\textbf{The skip-wrong penalty is essential to RL training.}
We isolate the contribution of the \emph{skip-wrong} reward term, a $-0.5$ penalty applied whenever the policy chose \texttt{skip} (bypassing the world model) and answered incorrectly. Without this term, a wrong \texttt{skip} costs nothing while a wrong \texttt{call\_wm} costs at least the per-action cost ($0.30$ for a typical $3$-step plan with $\text{action\_cost}{=}0.1$), which favors skipping under uncertainty. Reintroducing the penalty makes a wrong skip worse than a short wrong WM call (e.g., $-0.5$ versus $-0.3$ for three actions), restoring an incentive to query the world model. Table~\ref{tab:ablation_reward} reports the result. The skip-wrong term lifts overall test accuracy from $62.67\%$ to $77.33\%$ ($+14.66$ points), with the largest gains on \emph{action-consequence} ($+18.92$), \emph{ego-movement} ($+17.39$), and \emph{obj-movement} ($+17.39$); without it, the policy over-skips and its accuracy approaches the no-WM baseline. The broader penalty sweep is given in Appendix~\ref{app:penalty}.

\begin{figure*}[t]
  \centering
  \begin{subfigure}[t]{0.32\linewidth}
    \centering
    \includegraphics[width=\linewidth]{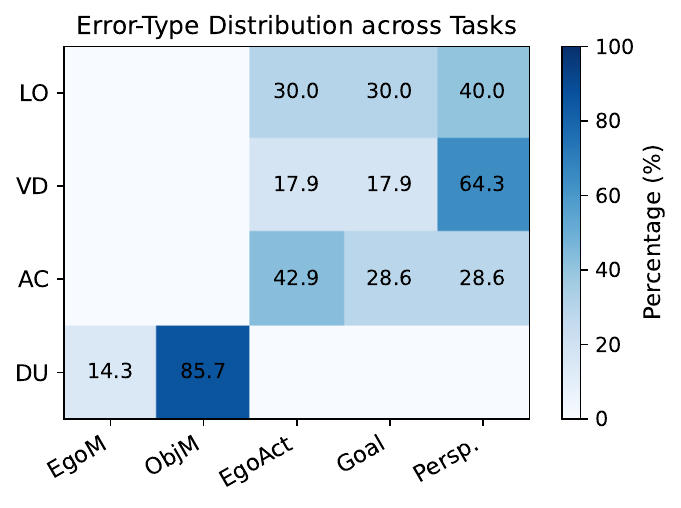}
    \caption{Error types vary across task categories.}
    \label{error-task}
  \end{subfigure}\hfill
  \begin{subfigure}[t]{0.32\linewidth}
    \centering
    \includegraphics[width=\linewidth]{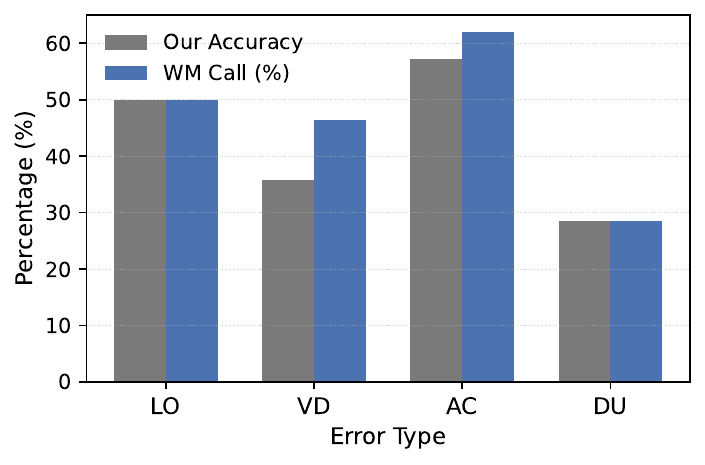}
    \caption{World-model usage and accuracy gains depend on error type.}
    \label{erro-wm}
  \end{subfigure}\hfill
  \begin{subfigure}[t]{0.32\linewidth}
    \centering
    \includegraphics[width=\linewidth]{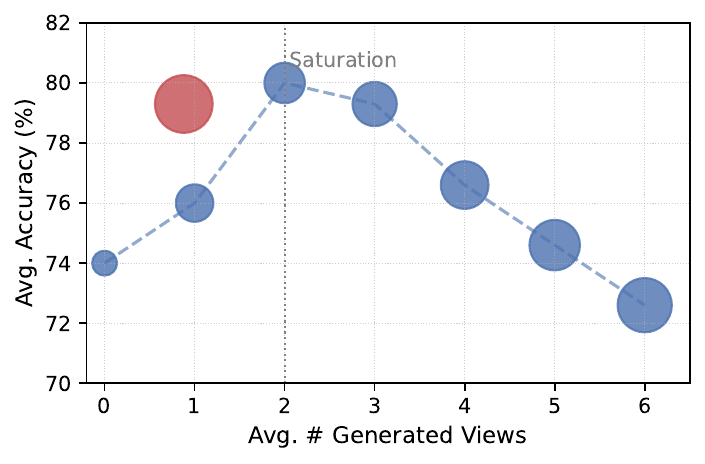}
    \caption{Accuracy saturates as more imagined views are generated.}
    \label{acc-view}
  \end{subfigure}
  \caption{Analysis of when and how much to invoke world-model imagination. 
  }
  \label{fig:wm_calling}
\end{figure*}

\subsection{When and How Much a World Model is Needed for Visual Spatial Reasoning?}

To diagnose \emph{when} world-model imagination is necessary, we manually classify MLLM failures on SAT-Real into four error types: \textbf{(1) Limited Observability} (LO, 15.2\%): required information occluded or out of view; \textbf{(2) Viewpoint Dependence} (VD, 42.4\%): answer depends on transforming between egocentric and object-centric frames; \textbf{(3) Action-Conditioned Reasoning} (AC, 31.8\%): answer depends on the scene state after a hypothetical action; and \textbf{Dynamics Understanding} ((4) DU, 10.6\%): temporal reasoning about camera or object motion.
\Cref{error-task} shows that SAT task categories do not map one-to-one with error types but exhibit \emph{compositional} patterns: EgoAct is dominated by AC errors (post-action viewpoints), Pers.\ by VD errors (reference-frame transformation), and ObjM by DU errors (temporal dynamics). LO errors appear across multiple categories, indicating that occlusion and limited field-of-view are general failure sources. This decomposition lets us study WM utility through error structure rather than surface task labels.

\begin{AIbox}{RQ1: When to call WM?}
WM should be used selectively, primarily when reasoning requires predicting future states under hypothetical actions rather than reinterpreting existing visual evidence.
\end{AIbox}

\paragraph{World models are most needed for action-conditioned reasoning.}
\Cref{erro-wm} shows that WM imagination yields the largest gain (\textbf{+57.1\%}) on AC errors, where the answer depends on the post-action scene state, e.g., counterfactual queries like ``what if I turn left by 90°?''. By contrast, DU errors require only reference-frame transformation over the current view and benefit much less (+28.5\%). LO and VD errors fall between these extremes: rendering can reveal occluded content (LO) or visualize the scene from a different vantage (VD), but adds little when the transformation can be inferred symbolically from the original observation. These findings indicate that WM utility is highly instance-dependent. It is most useful when the question references a future or counterfactual scene, often unnecessary for static reinterpretation of what is already observed.

\begin{AIbox}{RQ2: How much imagination is needed?}
Visual spatial reasoning benefits from \emph{targeted} rather than extensive WM imagination.
\end{AIbox}

\paragraph{Spatial reasoning requires limited imagination.}
\Cref{acc-view} sweeps fixed-budget baselines with a predetermined number of imagined views. Even a single targeted view raises accuracy by ~2\% over the no-imagination baseline (74.0\%~$\rightarrow$~76\%), and a second view captures most of the remaining headroom (76\%~$\rightarrow$~80\%). Beyond two views, additional rollouts bring no further gains and eventually degrade performance, as accumulated rendering artifacts and redundant content begin to confuse the downstream reasoner. Targeted, low-budget imagination is what spatial reasoning needs, while exhaustive scaling is wasteful at best and harmful at worst.

\begin{figure*}
    \centering
    \includegraphics[width=0.98\linewidth]{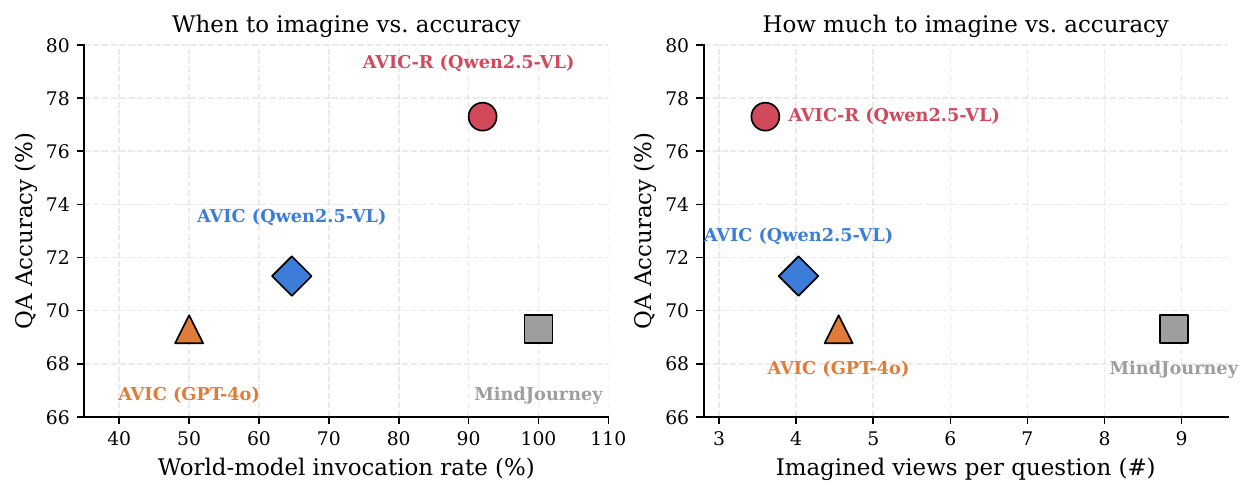}
    \caption{\textbf{Adaptive imagination achieves higher accuracy at lower cost on SAT-Real.} \emph{Left:} world-model invocation rate (\emph{when} to imagine) vs.\ QA accuracy. \emph{Right:} imagined views per question (\emph{how much}) vs.\ QA accuracy. \modelR~achieves the highest accuracy (\textbf{77.3\%}) using the fewest imagined views (\textbf{3.60} vs.\ 8.90 for always-on).}

    \label{fig:avicr}
\end{figure*}

\paragraph{\modelR~learns better when and how much to imagine.}
The findings above point to a simple rule for visual test time scaling with WM: we should call WM for visual spatial reasoning mainly on action-conditioned questions, and use only targeted views per call. 
\Cref{fig:avicr} (and Appendix~\Cref{tab:wm-call-budget}) shows that \modelR~learns both parts, while alternative policies do not.
On the \emph{when} axis, \modelR~calls WM on 92\% of questions on average, with strong differences across categories: 100\% on EgoAct (dominated by AC errors) versus 78.8\% on Pers.(dominated by VD errors). 
This category-aware behavior emerges from QA-correctness and WM-cost rewards alone, and no per-category labels are provided. Other alternatives are far less stable: \model~with GPT-4o calls 100\% on EgoM but 0\% on ObjM, and zero-shot Qwen2.5VL over-skips at 64.7\%, ending 6 points behind \modelR~(71.3\% vs.\ 77.3\%). 
Calling WM more often is not the point, but calling it on the right questions is.
On the \emph{how-much} axis, \modelR~uses 3.60 views per question, fewer than every selective baseline (zero-shot Qwen 4.03, GPT-4o 4.55) and less than half of MindJourney's 8.90, while still reaching the best accuracy. 
In short, both behaviors emerge from our lightweight RL scheme: the policy learns when to imagine and how much to imagine, just from QA correctness and a WM cost penalty.

\vspace{-1mm}
\section{Conclusion}
\vspace{-1mm}
In this paper, we study visual spatial reasoning with world models through the lens of adaptive test-time scaling, finding that always-on imagination is often unnecessary and even misleading. We introduce \model, which selectively decides \emph{when} and \emph{how much} to imagine at inference time, and \modelR, which trains this gating policy end-to-end via lightweight RL from QA-correctness and WM-cost signals. Across spatial reasoning and embodied navigation benchmarks, our framework achieves competitive or state-of-the-art results while substantially reducing world-model calls, tokens, and inference time; notably, \modelR~with a small open-source policy outperforms pipelines that use proprietary backbones as the policy. Our analysis shows world-model imagination is most beneficial for action-conditioned reasoning and requires only limited, targeted views, highlighting the importance of instance-dependent TTS for efficient and reliable reasoning with world models.

\textbf{Limitations \& Broader Impacts.} See Appendix for limitations and broader impacts discussion.

\section{Acknowledgments}
This work was supported by ONR Grant N00014-23-1-2356, ARO Award W911NF2110220, DARPA ECOLE Program No. HR00112390060, Microsoft Accelerate Foundation Models Research (AFMR) grant program, NSF AI Engage Institute DRL-2112635, and Cisco and Capital One Faculty Awards. The views contained in this article are those of the authors and not of the funding agency.

\bibliographystyle{plain}
\bibliography{main}

\newpage
\appendix

\section{Implementation Details}\label{app:implementation}

\paragraph{\modelR Training Details.}
We post-train Qwen2.5-VL-7B-Instruct with LoRA~\cite{hu2022lora} ($r{=}8$, $\alpha{=}16$, dropout $0.05$, applied to the \texttt{q/k/v/o} projections; ${\sim}5$M trainable parameters out of $8.3$B) using online GRPO~\cite{shao2024deepseekmath}. For each question we sample $K{=}16$ rollouts from the policy with temperature $1.0$, $\text{top-}p{=}0.95$, $\text{top-}k{=}50$, and a $512$-token response budget; advantages are computed by per-question (group) reward normalization. The PPO surrogate uses a clip ratio of $\epsilon{=}0.2$ and a K3 KL penalty to the frozen base policy with $\beta{=}0.1$, where the same backbone with the LoRA adapter \emph{disabled} serves as the reference policy (no separate frozen copy). 
Each rollout's reward combines task correctness, computed by matching the frozen GPT-4o answer to the dataset ground-truth label, with an action cost of $0.1$ per atomic step, a parse-failure penalty of $-0.5$, and an additional $-0.5$ \emph{skip-wrong} penalty when the policy bypasses the world model and answers incorrectly; this last term is essential to prevent the policy from collapsing to ``always skip''. The action vocabulary consists of $9^{\circ}$ rotations and $0.25$\,m forward steps, capped at $6$ atomic actions per plan. We optimize with AdamW, $\text{lr}{=}2{\times}10^{-5}$, weight decay $0$, gradient clipping at $1.0$, per-device batch size $1$ and no gradient accumulation, yielding an effective batch of $8$ questions per optimizer step on $8{\times}$A100-80\,GB GPUs (DDP via \texttt{torchrun}). 
The world model is Stable Virtual Camera~\cite{zhou2025stable} run in \texttt{img2trajvid\_s-prob} mode (CFG $4.0$, $8$ target views, trajectory prior, \texttt{interp} chunking, short side $576$). Training data is a curated $30/70$ mix of GPT-4o-prescored \emph{easy-skip} and \emph{needs-WM} questions; we train for up to $300$ optimizer steps and select the checkpoint that achieves the best held-out accuracy.
We train \modelR using GPT-4o as the frozen QA model and apply the resulting policy to other QA backbones without retraining (Table~\ref{table:SAT}).

\paragraph{Training compute and accounting.}
Training uses eight A100-80\,GB GPUs for up to 300 optimizer steps, with checkpoint selection based on held-out accuracy. Only the approximately 5M LoRA parameters are updated. World-model generation and frozen-QA queries supply the online rewards; the trajectory verifier is absent from training and is used only during inference. Parameter-efficient updates do not remove the cost of these online queries. Inference timing is reported separately in Table~\ref{tab:efficiency}; that table does not isolate verifier latency as a separate column and should not be read as a training-cost measurement.

\paragraph{Balanced data curation.} Random training sampling is inefficient: as shown in~\cref{empirical statistics}, most SAT questions are already answered correctly without imagination (Case~3), so every reasonable rollout produces $r\!\approx\!+1$, the within-group variance vanishes, and~\cref{eq:advantage} yields zero gradient. The questions that drive learning are those the base VLM fails on but a targeted imagined view would correct. We therefore pre-score a candidate pool with a strong reference QA model under the $\texttt{skip}$ policy and split each question into \emph{easy-skip} (base correct) or \emph{needs-imagination} (base wrong). The final training set (\~ 3000 examples) retains every needs-imagination instance and sub-samples easy-skip ones at a 30\% ratio, balancing learning signal against an anchor that prevents collapse onto always-call-WM.

\paragraph{Lenient schema parsing.} High-temperature sampling occasionally yields JSON outputs that are syntactically broken but semantically recoverable: $d$ set to an action verb (\texttt{turn-left}) rather than a meta-decision, or $\pi$ written as free-form strings (\texttt{"turn-right 9 degrees"}). A strict parser would reject these as parse failures, conflating format errors with semantic confusion. We instead use a lenient parser with a salvage stage that infers $d=\texttt{call\_wm}$ whenever $\pi$ is non-empty regardless of the literal $d$ value, and parses natural-language action strings into structured records via regex. Truly unrecoverable outputs still incur $r_{\text{parse}}$. The finer-grained reward distinguishes ``intent recovered, format mangled'' (taught via the call-WM reward) from ``no plan at all'' (parse failure), supplying GRPO with a smoother gradient.

\textbf{Metrics.} We evaluate SAT/MMSI spatial reasoning benchmark with multiple-choice QA accuracy. 
In the R2R navigation setting, the agent must follow natural language instructions in indoor environments. We integrate our adaptive visual test-time scaling framework into the navigation pipeline and measure performance. 
We evaluate navigation performance using four standard metrics: Navigation Error (NE), Oracle Success Rate (OSR), Success Rate (SR), and Success weighted by Path Length (SPL). NE measures the geodesic distance between the agent's final position and the target, while SR reports the fraction of episodes where the final position is within a predefined success threshold. OSR measures whether the agent ever reaches within the success threshold at any point along its trajectory, reflecting exploration ability independent of stopping. SPL jointly evaluates success and efficiency by weighting successful episodes by the ratio between shortest-path length and the actual trajectory length.

\textbf{Prompts.} We provide extra technical details of our adaptive visual test time scaling framework.
In~\cref{tab:score_prompt}, we provide verifier prompts that are used to score each generated trajectory, and in~\cref{tab:policy}, we provide prompts for world model gating and action planning in our policy model.

\section{Extra Experiments}\label{app:extra}

\begin{figure*}[]
    \centering
    \includegraphics[width=1\linewidth]{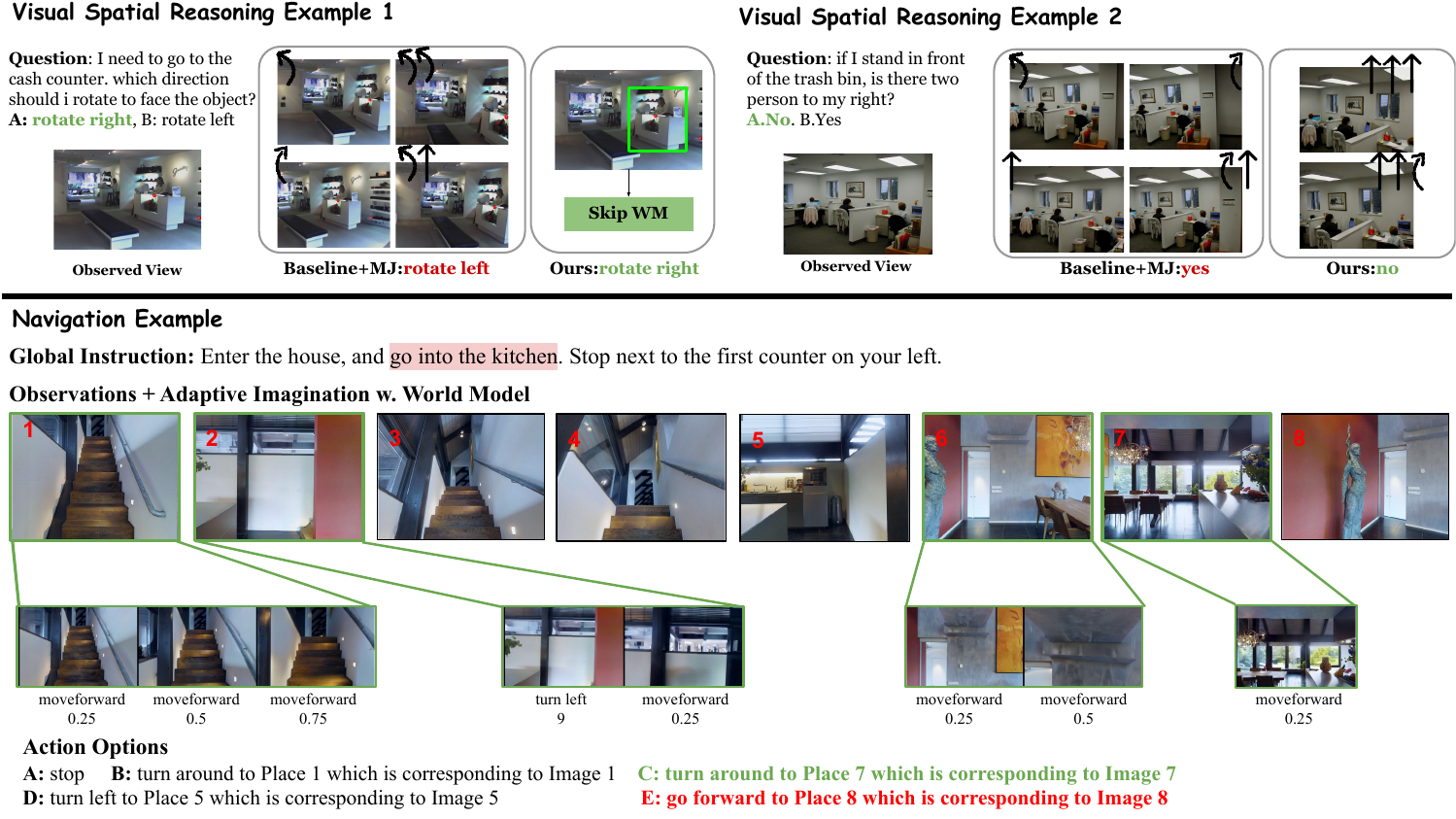}
    \caption{Qualitative examples on SAT of the always-on imagination method and our adaptive method, as well as the R2R navigation task. In the navigation example, the green option is selected by the model with adaptive imagination via our method, while the red one is without world model imagination. 
    }
    \label{fig:vis}
\end{figure*}

\paragraph{Qualitative Analysis.} We also provide qualitative examples as illustrated at the top of~\cref{fig:vis}. We compare our adaptive visual TTS method with the always-on imagination method, MindJourney (MJ). 
In the first example, the target object (the \emph{cash counter}) is already clearly visible in the observed view.
Our method correctly identifies that additional visual imagination is unnecessary and directly skips world model. In contrast, MJ indiscriminately invokes the world model, generating multiple imagined views that introduce misleading evidence and ultimately lead to an incorrect prediction. In the second example, \model \  yields the correct answer by selectively imagining the state where the agent is in front of the \emph{trash bin}. In contrast, MJ performs dense imagination and generates views that do not accurately reflect this critical spatial condition, leading to an incorrect prediction. Furthermore, we present a qualitative navigation example at the bottom of~\cref{fig:vis}. Our adaptive visual test-time scaling selectively augments informative indoor observations (e.g., zooming in or turning to explore nearby views), enabling the agent to better inspect the environment and align its actions with the global instruction (\emph{``go to the kitchen''}). 
In contrast, the baseline without visual imagination lacks sufficient perceptual evidence and consequently chooses an incorrect direction.

\paragraph{Sensitivity to Errors from World Models.}
 Our work is motivated by the observation that imperfect imagination can introduce misleading or noisy evidence that may degrade performance (Sec.~\ref{sec:analysis}, Figs.~1--2). To further evaluate robustness, we conduct additional experiments using \textbf{Cosmos}~\cite{ali2025world} in~\cref{tab:wm-robust} as an alternative world model, which tends to produce less visually stable camera trajectories and noisier geometric structures compared to SVC.

 \paragraph{Stage-wise Computation Cost.}
The decision module adds a lightweight inference step whose cost is small relative to world-model (WM) generation. Rather than always invoking expensive imagination, it predicts \emph{when} and \emph{how} to use the WM, reducing unnecessary calls. The table shows that policy cost is minor compared to WM cost and is offset by large savings. Although \model~introduces $\sim$14.9\,s of policy overhead, it reduces WM time by $\sim$153.7\,s ($163.32 \rightarrow 9.59$), yielding a \textbf{$\sim$6$\times$ reduction in total time} ($177.84 \rightarrow 29.04$) and \textbf{$\sim$20$\times$ fewer tokens} ($162.6 \rightarrow 7.6$\,k), while improving accuracy by 2.0 points over always-on.  \model~achieves both higher accuracy and substantially lower overall computation.

\begin{table*}[h]
\centering
\small
\setlength{\tabcolsep}{4pt}
\caption{\textbf{Inference cost and accuracy on SAT-Real.} \model~adds modest policy-inference overhead but substantially reduces world-model time and token usage, achieving the highest accuracy at $\sim$6$\times$ lower total time and $\sim$20$\times$ fewer tokens than always-on imagination.}
\label{tab:efficiency}
\begin{tabular}{lrrrrrrr}
\toprule
\textbf{Method} & \textbf{Policy (s)} & \textbf{WM (s)} & \textbf{QA (s)} & \textbf{Total (s)} & \textbf{Tokens (k)} & \textbf{WM Calls} & \textbf{Acc. (\%)} \\
\midrule
Baseline (no WM)         & 0     & 0      & 3.9   & 3.9    & 0.7   & 0     & 74.0 \\
MindJourney (always-on)  & 0     & 163.32 & 14.52 & 177.84 & 162.6 & 12.42 & 77.3 \\
\model                   & 14.92 & 9.59   & 4.53  & 29.04  & 7.6   & 0.73  & 79.3 \\
\bottomrule
\end{tabular}
\end{table*}

\paragraph{Robustness across runs.}
To assess the stability of \modelR, we repeat our main evaluation three times with independently sampled rollouts; all other hyperparameters are held fixed. The overall accuracy gave a mean of $69.33$ with a sample standard deviation of $0.77$ (standard error of the mean $0.44$).

\begin{table}[]
\centering
\small
\setlength{\tabcolsep}{6pt}
\caption{\textbf{Analysis of \emph{when} and \emph{how much} to imagine on SAT-Real.} We compare four imagination policies along three axes: fraction of questions on which the world model is invoked (\emph{top}), average number of imagined views per question (\emph{middle}), and resulting QA accuracy (\emph{bottom}).}
\begin{tabular}{l ccccc | r}
\toprule
\textbf{Method} & \multicolumn{1}{c}{\textbf{EgoM}} & \multicolumn{1}{c}{\textbf{ObjM}} & \multicolumn{1}{c}{\textbf{EgoAct}} & \multicolumn{1}{c}{\textbf{Goal}} & \multicolumn{1}{c}{\textbf{Pers}} & \multicolumn{1}{c}{\textbf{Avg.}} \\
\midrule
\rowcolor{blue!10}
\multicolumn{7}{c}{\textit{when to call world model (call\_wm \%)}} \\
MindJourney (always-on)  & 100.0 & 100.0 &       100.0     & 100.0 & 100.0 & 100.0  \\
\model~(GPT-4o)   & 100.0 &   0.0          &       62.2     &       32.4     &       54.5     &       50.0     \\
\model~(Qwen2.5VL)              &       47.8     &       60.9     &       78.4     &       73.5     &       54.5     &       64.7     \\
\modelR (Qwen2.5VL)               &       91.3     &       91.3     & 100.0 &       97.1     &       78.8     &       92.0     \\
\midrule
\rowcolor{blue!10}
\multicolumn{7}{c}{\textit{how much to imagine (\# img per question)}} \\

MindJourney (always-on)   & 11.13 & 4.83  & 8.37  & 11.03 & 8.55  & 8.90  \\
\model~(GPT-4o)     &        3.30    &  0.00            &        5.30    &        4.18    &        5.39    &        4.55    \\
\model~(Qwen2.5VL)             &        4.09    &        3.07    &        4.90    &        3.28    &        4.39    &        4.03    \\
\modelR (Qwen2.5VL)            &        2.95    &        3.00    &        4.59    &        2.79    &        4.23    &        3.60    \\
\midrule
\rowcolor{blue!10}
\multicolumn{7}{c}{\textit{task accuracy (\%)}} \\
MindJourney (always-on)   &       78.3     &       60.9     &       78.4     &       70.6     &       57.5     &       69.3    \\
\model~(GPT-4o)    & \textbf{87.0}  &       69.6     &       64.9     &       82.4     &       48.5     &       69.3   \\

\model~(Qwen2.5VL)              &       65.2     &       73.9     &       64.9     & \textbf{91.2}  & \textbf{60.6}  &       71.3    \\
\modelR (Qwen2.5VL) &       82.6     & \textbf{82.6}  & \textbf{81.1}  & \textbf{91.2}  &       51.5     & \textbf{77.3} \\
\bottomrule
\end{tabular}

\label{tab:wm-call-budget}
\end{table}

The Cosmos result remains above the o1 baseline ($77.3\%$ versus $74.6\%$), but the gain is smaller than with SVC ($85.3\%$). Thus the method remains sensitive to world-model quality. Appendix~\ref{app:transfer} extends this comparison to the RL-trained policy without retraining, and Appendix~\ref{app:interpretation} discusses why these results do not isolate the causes of misleading imagined evidence.

\paragraph{Evaluations on Additional Benchmark.}
We extend our evaluation to \textbf{MindCube}~\cite{yin2025spatial} in~\cref{tab:mindcube}. \model~improves performance on this fine-grained spatial reasoning benchmark, indicating that our method generalizes across tasks.

\textbf{Framework Error Analysis.}
Table~\ref{tab:recall_precision} reports outcome-based recall and precision for the zero-shot GPT-4o policy. These metrics depend on the generated plans, the rendered evidence, and the final reasoner as well as the gating decision. The controlled Qwen2.5-VL-7B comparison in Appendix~\ref{app:penalty} distinguishes zero-shot from RL-trained behavior. Invocation rates and view budgets are reported separately in Table~\ref{tab:wm-call-budget}.

\begin{table}[h]
\centering
\small

\begin{minipage}[t]{0.40\linewidth}
    \vspace{0pt}
    \centering
    \setlength{\tabcolsep}{4pt}

    \caption{Results on MindCube-Tiny}
    \label{tab:mindcube}
    \begin{tabular}{lr}
        \toprule
        \textbf{Method} & \textbf{Acc.\ (\%)} \\
        \midrule
        GPT-4o          & 36.5 \\
        GPT-4o + \model & \textbf{38.7}~$(+2.2)$ \\
        \bottomrule
    \end{tabular}

    \par\medskip

    \caption{Performance across world models. Based on o1.}
    \label{tab:wm-robust}
    \begin{tabular}{lr}
        \toprule
        \textbf{Method} & \textbf{Acc.\ (\%)} \\
        \midrule
        o1                  & 74.6 \\
        o1 + \model~(SVC)    & 85.3 \\
        o1 + \model~(Cosmos) & 77.3 \\
        \bottomrule
    \end{tabular}
\end{minipage}
\hfill

\begin{minipage}[t]{0.56\linewidth}
    \vspace{0pt}
    \centering
    \setlength{\tabcolsep}{4pt}

    \caption{\textbf{Gating recall and precision} (\%) per category
    for the zero-shot GPT-4o policy in \model,
    rather than the RL-trained \modelR policy.
    \emph{Recall}: fraction of WM-needed questions on which the policy calls WM.
    \emph{Precision}: fraction of WM-called questions that actually benefit from WM.}
    \label{tab:recall_precision}

    \resizebox{\linewidth}{!}{%
        \begin{tabular}{lcccccc}
            \toprule
            \textbf{Metric} & \textbf{EgoM} & \textbf{ObjM}
            & \textbf{EgoAct} & \textbf{Goal} & \textbf{Pers.}
            & \textbf{Avg.} \\
            \midrule
            Recall    & 100.0 & 33.3 & 55.6 & 33.3 & 52.6 & 43.9 \\
            Precision &   5.6 & 50.0 & 19.2 & 22.2 & 62.5 & 27.1 \\
            \bottomrule
        \end{tabular}%
    }
\end{minipage}

\end{table}

\begingroup
\section{Additional Ablations and Analysis}
\label{app:additional-analysis}

We further examine the contributions of gating, planning, and trajectory selection; the wrong-skip penalty; transfer across world models and reasoners; and performance on SAT-Synthesized. We distinguish the configurations used in these experiments because the policy and reasoner need not be the same model.

\subsection{Contributions of Gating, Planning, and Verification}
\label{app:modules}

Table~\ref{tab:additional-modules} evaluates a Qwen2.5-VL-7B policy with a GPT-4o reasoner on SAT-Real. Gating and planning are produced jointly in one policy forward pass. Consequently, the gating-only condition retains the policy's gate but substitutes the always-on baseline's greedy search whenever imagination is requested. In the gating-and-planning conditions without the MLLM verifier, we answer each rollout separately and aggregate answers by majority vote. This \emph{answer-level} self-consistency is distinct from majority voting over the policy's \emph{gating decisions} in the full framework.

\begin{table}[htbp]
\centering
\small
\setlength{\tabcolsep}{5pt}
\caption{Component ablation on SAT-Real with a Qwen2.5-VL-7B policy and GPT-4o reasoner. Calls and views are averages per question. ``Plan'' denotes policy-generated plans; the gating-only variants instead use baseline greedy search. For the gating-and-planning variants, removing the verifier substitutes answer-level self-consistency.}
\label{tab:additional-modules}
\begin{tabular}{lcccrrr}
\toprule
Method & Gate & Plan & Verify & WM calls & Views & Acc. (\%) \\
\midrule
No imagination & -- & -- & -- & 0 & 0 & 60.3 \\
Always-on & -- & -- & -- & 12.34 & 8.90 & 69.3 \\
\model{} & $\checkmark$ & -- & -- & 7.40 & 5.34 & 68.0 \\
\model{} & $\checkmark$ & $\checkmark$ & -- & 2.21 & 4.93 & 69.3 \\
\model{} & $\checkmark$ & $\checkmark$ & $\checkmark$ & 1.81 & 4.03 & 71.3 \\
\midrule
\modelR & $\checkmark$ & -- & -- & 7.94 & 5.73 & 72.6 \\
\modelR & $\checkmark$ & $\checkmark$ & -- & 3.37 & 4.49 & 75.3 \\
\modelR & $\checkmark$ & $\checkmark$ & $\checkmark$ & 3.03 & 3.60 & \textbf{77.3} \\
\bottomrule
\end{tabular}
\end{table}

For the zero-shot policy, gating alone reduces computation but does not retain all the accuracy of always-on imagination. Adding policy-generated plans raises zero-shot accuracy from $68.0\%$ to $69.3\%$ while reducing average calls from $7.40$ to $2.21$. Replacing answer voting with the verifier adds $2.0$ percentage points for both zero-shot and RL-trained policies. RL adds $6.0$ points in each gating-and-planning configuration. The full \modelR achieves $77.3\%$, compared with $69.3\%$ for always-on imagination, with $3.03$ rather than $12.34$ calls. These are system-level comparisons: their computation budgets also differ, so they do not isolate a component's effect at matched compute.

\subsection{Effect of the Trajectory Selection Rule}
\label{app:selection}

We additionally compare selection rules using GPT-4o as both policy and reasoner with $M=5$ policy samples (Table~\ref{tab:additional-selection}). Random selection chooses a candidate trajectory; answer self-consistency votes over the answers obtained from the candidates; and the MLLM verifier scores each complete rendered trajectory before selecting the highest-scoring one. The verifier prompt is provided in Table~\ref{tab:score_prompt}. The candidate-set oracle counts a question as successful if at least one of the five candidates yields the correct answer.

\begin{table}[htbp]
\centering
\small
\caption{Downstream QA accuracy under different selection rules on SAT-Real. Both the policy and reasoner are GPT-4o, with $M=5$. These scores measure answer accuracy, not the verifier's trajectory-ranking accuracy.}
\label{tab:additional-selection}
\begin{tabular}{lr}
\toprule
Selection rule & Acc. (\%) \\
\midrule
Random & 70.0 \\
Answer self-consistency & 75.3 \\
MLLM verifier & 77.3 \\
Candidate-set oracle & 88.0 \\
\bottomrule
\end{tabular}
\end{table}

The verifier improves accuracy by $7.3$ points over random selection and $2.0$ points over answer self-consistency, recovering approximately $41\%$ of the random-to-oracle gap. Substantial selection headroom remains. 
This separate selection experiment uses a GPT-4o policy; its $77.3\%$ result should not be conflated with the numerically equal \modelR result in Table~\ref{tab:additional-modules}.

\subsection{Wrong-Skip Penalty and Outcome-Based Precision}
\label{app:penalty}

Table~\ref{tab:additional-penalty} varies the wrong-skip penalty $\beta_s$ using the same Qwen2.5-VL-7B policy backbone. With action cost $c=0.1$, the setting $\beta_s=0.3$ equalizes the penalties for a wrong skip and a wrong call with a three-action plan. It is not symmetric for every plan length, because the call cost is $c|\pi|$.

\begin{table}[htbp]
\centering
\small
\caption{Wrong-skip penalty sweep on SAT-Real with a Qwen2.5-VL-7B policy. Precision and accuracy are percentages. The zero-shot row is evaluated without RL; the remaining rows use RL with the specified penalty.}
\label{tab:additional-penalty}
\begin{tabular}{lrr}
\toprule
Policy / $\beta_s$ & Precision & Acc. \\
\midrule
Zero-shot & 38.2 & 71.3 \\
RL, $0$ & 19.5 & 62.67 \\
RL, $0.3$ & 43.2 & 74.6 \\
RL, $0.5$ & \textbf{50.5} & \textbf{77.3} \\
RL, $1.0$ & 36.1 & 67.2 \\
\bottomrule
\end{tabular}
\end{table}

The default $\beta_s=0.5$ achieves the highest precision and accuracy among the tested settings. Both removing the penalty and increasing it to $1.0$ degrade performance; low precision therefore cannot be attributed to reward asymmetry alone. Here, precision is outcome-based: a call is counted as beneficial only when it improves the final answer. It consequently depends on planning, rendering, and evidence integration as well as the gate. The $27.1\%$ precision in Table~\ref{tab:recall_precision} belongs to the zero-shot GPT-4o policy, whereas the $38.2\%$ and $50.5\%$ values here belong to zero-shot and RL-trained Qwen policies, respectively.

\subsection{Transfer across World Models and Reasoners}
\label{app:transfer}

Table~\ref{tab:additional-transfer} extends the world-model comparison using o1 as the reasoner. The zero-shot \model{} policy has not been trained on SVC-generated views, so its SVC--Cosmos gap is not evidence of policy-training overfitting to SVC. For \modelR, we replace SVC with Cosmos at inference without retraining the policy originally optimized using SVC and the frozen GPT-4o QA model.

\begin{table}[htbp]
\centering
\small
\setlength{\tabcolsep}{5pt}
\caption{Transfer across world models on SAT-Real with the o1 reasoner. The zero-shot \model{} policy is o1. The \modelR policy is Qwen2.5-VL-7B, trained using GPT-4o and SVC, then evaluated with either world model without retraining. Gains are percentage points over no imagination.}
\label{tab:additional-transfer}
\begin{tabular}{llrr}
\toprule
Policy & World model & Acc. (\%) & Gain \\
\midrule
No imagination & -- & 74.6 & -- \\
\model{} (zero-shot) & SVC & 85.3 & $+10.7$ \\
\model{} (zero-shot) & Cosmos & 77.3 & $+2.7$ \\
\modelR & SVC & 80.0 & $+5.4$ \\
\modelR & Cosmos & 76.0 & $+1.4$ \\
\bottomrule
\end{tabular}
\end{table}

Both policies improve over the baseline with either world model, but gains are smaller with Cosmos. This supports transfer in the tested setting while showing that the benefit remains sensitive to the renderer; it does not establish invariance to world-model quality.

The correctness reward compares the frozen QA model's answer with the dataset label. The QA model is therefore an answer generator, not a judge that defines correctness. Its reasoning biases can nevertheless affect which trajectories receive reward. Table~\ref{table:SAT} evaluates the same trained \modelR policy without retraining across four reasoners: gains over their respective baselines are $10.0$ points for InternVL3-14B, $17.0$ for GPT-4o, $6.0$ for GPT-4.1, and $5.4$ for o1. These results demonstrate transfer to the evaluated reasoners, while the largest gain remains on the reasoner used during training.

\subsection{Conditional Oracle Analysis}
\label{app:oracles}

We examine two privileged interventions on \modelR (Table~\ref{tab:additional-oracles}). Oracle gating selects skip versus call using precomputed ground-truth information while retaining the policy's plans and trajectories. The second intervention additionally selects the best trajectory among the sampled candidates.

\begin{table}[htbp]
\centering
\small
\caption{Conditional oracle analysis for \modelR on SAT-Real. These interventions use ground-truth information unavailable at deployment and retain the sampled candidate set.}
\label{tab:additional-oracles}
\begin{tabular}{lr}
\toprule
Setting & Acc. (\%) \\
\midrule
\modelR & 77.3 \\
With oracle gating & 88.0 \\
With oracle gating and trajectory selection & 95.3 \\
\bottomrule
\end{tabular}
\end{table}

The successive gains of $10.7$ and $7.3$ points quantify conditional headroom under privileged decisions, not a unique causal decomposition or a model-independent upper bound. They do not establish reasoner reliability on arbitrary imagined evidence. The $75.3\%$ oracle in Section~\ref{sec:analysis} is restricted to MindJourney's fixed pipeline. The $88.0\%$ selection oracle in Table~\ref{tab:additional-selection} also uses a different policy configuration from the oracle-gating experiment here.

\subsection{Additional Evaluation on SAT-Synthesized}
\label{app:synth}

Table~\ref{tab:additional-synth} reports the additional SAT-Synthesized comparison. \model{} and \modelR improve accuracy over MindJourney by $1.5$ and $4.8$ points, respectively, while using fewer world-model calls. This extends the observed accuracy--computation trade-off beyond SAT-Real.

\begin{table}[htbp]
\centering
\small
\caption{Additional results on SAT-Synthesized. Average WM calls are reported per question.}
\label{tab:additional-synth}
\begin{tabular}{lrr}
\toprule
Method & Acc. (\%) & Avg. WM calls \\
\midrule
GPT-4o, no imagination & 61.0 & 0 \\
MindJourney & 72.3 & 12.34 \\
\model{} & 73.8 & 2.34 \\
\modelR & \textbf{77.1} & 3.92 \\
\bottomrule
\end{tabular}
\end{table}

\subsection{Interpretation and Scope}
\label{app:interpretation}

\paragraph{Invocation frequency and imagination budget.}
The $54\%$ ``unnecessary'' fraction in Section~\ref{sec:analysis} diagnoses the original fixed-imagination setting; it is not a target rate for the learned \modelR gate, which invokes the WM on $92\%$ of questions (Table~\ref{tab:wm-call-budget}). Its $50.5\%$ outcome-based precision still leaves substantial room for improvement. Frequent invocation can nevertheless coexist with fewer calls and views per question when each call uses a shorter plan than exhaustive exploration.

\paragraph{Misleading imagined evidence.}
Case~2 is outcome-based. Rendering errors, plausible but uninformative trajectories, and failures to integrate imagined evidence can all contribute. The world-model comparison supports sensitivity to rendering quality but does not quantitatively separate these causes.

\paragraph{Imagination and navigation actions.}
In R2R, MapGPT's graph-defined navigation candidates remain unchanged. The $9^{\circ}$ rotations and $0.25$\,m forward motions augment currently observed panoramas through virtual camera motion; they neither move the agent nor require hypothetical graph nodes. The discrete, six-action interface (Table~\ref{tab:policy}) limits imagination resolution and horizon; continuous control and richer interactions remain outside this evaluation.

\paragraph{Statistical scope.}
The additional SAT-Real ablations use $150$ questions and report point estimates without repeated-run variability or confidence intervals. Small differences should be interpreted cautiously; these results do not establish statistical significance or robustness across training seeds.
\par
\endgroup

\section{Impact Statement}
This work studies world-model-based visual imagination in visual spatial reasoning and highlights the limitations of existing always-on test-time imagination methods. Through systematic analysis, we show that indiscriminate visual imagination can be computationally inefficient and, in some cases, harmful due to misleading or redundant imagined views. Our findings emphasize the importance of adaptive test-time computation, demonstrating that effective spatial reasoning requires selectively invoking visual imagination only when necessary and scaling it appropriately.
Beyond the specific benchmarks studied, our insights are broadly applicable to multimodal agents that rely on test-time simulation, including embodied AI and interactive systems.

\section{Limitations}\label{app:limitations}
Our work focuses on adaptive imagination control for visual spatial reasoning and short-horizon embodied navigation; extending the framework to longer-horizon decision-making, manipulation, and broader multi-modal tasks is a natural next direction. The framework also assumes access to a separate visual world model and a fixed discrete action space, leaving room for richer extensions such as continuous action spaces, multiple specialized world models, or joint training of the gating policy with the world model itself. Incorporating world-model uncertainty into the reward signal is another promising direction for improving robustness under noisier imagined rollouts. Appendix~\ref{app:interpretation} further discusses the outcome-based failure taxonomy, the distinction between call frequency and imagination budget, and the statistical scope of the additional experiments.

\section{License}
We will make our code and models publicly accessible. 
We use standard licenses from the community and provide the following links to the licenses for the datasets, codes, and models that we used in this paper. 
For further information, please refer to the specific link.

\noindent\textbf{QWen2.5VL~\cite{bai2025qwen2}:} \href{https://huggingface.co/Qwen/Qwen2.5-VL-7B-Instruct}{Apache-2.0}

\noindent\textbf{SAT~\cite{ray2024sat}:} \href{https://huggingface.co/datasets/array/SAT}{MIT}

\noindent\textbf{MMSI~\cite{yang2025mmsi}:} \href{https://huggingface.co/datasets/RunsenXu/MMSI-Bench}{CC-BY-4.0}

\noindent\textbf{R2R~\cite{anderson2018vision}:} \href{https://bringmeaspoon.org/}{MIT}

\begin{table*}[t!]
\centering
\begin{minipage}{1\columnwidth}
    \centering
    \caption{Verifier prompts for scoring imagined view plans.}
    \begin{tcolorbox} 
        \centering
        \hspace{-6mm}
        \begin{tabular}{p{0.99\columnwidth}}
        \hspace{1mm}
        \begin{minipage}{0.99\columnwidth}
        
\textbf{Role.} You are an \emph{independent evaluator} for visual spatial reasoning.

\vspace{0.5em}
\textbf{Input.} A multiple-choice question, answer options, the current observation image(s), and \emph{one} candidate action plan.  
The plan includes imagined views rendered by a world model.

\vspace{0.5em}
\textbf{Task.} Score how \emph{useful} the imagined views are for answering the question.

\vspace{0.3em}
\textbf{Score Range.} Integer from 0 (not helpful, irrelevant, or low quality) to 9 (highly helpful and informative).

\vspace{0.5em}
\textbf{Scoring Guidelines.}
\begin{itemize}
    \item Assign higher scores if the imagined views reveal missing evidence needed to answer the question (e.g., resolving occlusion or viewpoint ambiguity).
    \item Assign higher scores if the imagined views are sharp, coherent, and visually consistent.
    \item Assign lower scores if the views are redundant, uninformative, distorted, or unrelated to the question.
    \item If the original observations are already sufficient, most plans should receive low scores.
\end{itemize}

\vspace{0.5em}
\textbf{Rules.}
\begin{itemize}
    \item Do \textbf{not} answer the question.
    \item Output \textbf{only} a single integer between 0 and 9.
    \item Do not output any additional text.
\end{itemize}

\vspace{0.5em}
\textbf{Output Example.}
\begin{center}
\texttt{5}
\end{center}

        \end{minipage}
        \end{tabular}
    \end{tcolorbox}
    \vspace{-2mm}
    \label{tab:score_prompt}
\end{minipage}
\end{table*}

\begin{table*}[t!]
\centering
\begin{minipage}{1\columnwidth}   
    \centering
    \caption{Policy model prompts for world model gating and action planning.}
    \begin{tcolorbox} 
        \centering
        \hspace{-6mm}
        \begin{tabular}{p{0.99\columnwidth}}
        \hspace{1mm}
        \begin{minipage}{0.99\columnwidth}
        
\textbf{Role.} You are a \emph{policy model} for spatial reasoning in a 3D indoor environment. Your goal is to decide whether to invoke a world model (WM) and, if needed, plan actions that acquire the most informative imagined views.

\vspace{0.5em}
\textbf{Input.} One or more images, a multiple-choice question, and answer options.

\vspace{0.5em}
\textbf{Tasks.}
\begin{itemize}
    \item Decide whether to \texttt{SKIP} or \texttt{CALL} the world model.
    \item If \texttt{CALL}, generate a short action plan (1--6 actions) to gather additional visual evidence.
\end{itemize}

\vspace{0.5em}
\textbf{Action Space (Discrete, Fixed).}
\begin{itemize}
    \item \texttt{move-forward} 0.25 meters
    \item \texttt{turn-left} 9 degrees
    \item \texttt{turn-right} 9 degrees
\end{itemize}

\vspace{0.5em}
\textbf{Action Composition Guidelines.}
\begin{itemize}

\item Repeated turns approximate larger rotations (e.g., 2 turns $\approx$ 18°, 5 turns $\approx$ 45°, 10 turns $\approx$ 90°).
    \item When a question specifies a larger angle, approximate it using repeated 9° turns.
\end{itemize}

\vspace{0.5em}
\textbf{When to Call the World Model.}
\begin{itemize}
    \item The answer is not directly observable from the current view.
    \item The question depends on perspective, facing direction, rotation, or left/right relations.
    \item The question requires reasoning about motion or state changes over time.
\end{itemize}

\vspace{0.5em}
\textbf{Constraints.}
\begin{itemize}
    \item Do not generate cancelling or oscillating actions (e.g., left then right).
    \item If turning, choose a single direction and turn monotonically.
\end{itemize}

\vspace{0.5em}
\textbf{Output Format (JSON only).}
\begin{verbatim}
{
  "decision": "skip" | "call_wm",
  "reason": "<one sentence>",
  "actions": [
    {"type": "move-forward" | "turn-left" | "turn-right",
     "value": <number>}
  ]
}
\end{verbatim}

        \end{minipage}
        \end{tabular}
    \end{tcolorbox}
    \vspace{-2mm}
    \label{tab:policy}
\end{minipage}
\end{table*}


\end{document}